\documentclass[runningheads]{llncs}

\usepackage[T1]{fontenc}
\usepackage{amsmath}
\usepackage{amssymb}
\usepackage{graphicx}

\begin{document}

\title{Multi-modal Imaging Genomics Transformer: Attentive Integration of Imaging with Genomic Biomarkers for Schizophrenia Classification}

\titlerunning{Multi-modal Imaging Genomics Transformer}


\author{Nagur Shareef Shaik \inst{1} \inst{2} \and
Teja Krishna Cherukuri \inst{1} \inst{2} \and Vince D. Calhoun \inst{1} \inst{2} \and Dong Hye Ye\inst{1} \inst{2}}

\authorrunning{N. Shaik et al.}

\institute{Georgia State University \and Center for Translational Research in Neuroimaging and Data Science (TReNDS)
\email{\{nshaik3,tcherukuri1\}@student.gsu.edu}, \email{\{vcalhoun,dongye\}@gsu.edu}  \\ Atlanta GA 30303, USA}
\maketitle

\vspace{-20pt}

\begin{abstract}
Schizophrenia (SZ) is a severe brain disorder marked by diverse cognitive impairments, abnormalities in brain structure, function, and genetic factors. Its complex symptoms and overlap with other psychiatric conditions challenge traditional diagnostic methods, necessitating advanced systems to improve precision. Existing research studies have mostly focused on imaging data, such as structural and functional MRI, for SZ diagnosis. There has been less focus on the integration of genomic features despite their potential in identifying heritable SZ traits. In this study, we introduce a Multi-modal Imaging Genomics Transformer (MIGTrans), that attentively integrates genomics with structural and functional imaging data to capture SZ-related neuroanatomical and connectome abnormalities. MIGTrans demonstrated improved SZ classification performance with an accuracy of 86.05\% ($\pm$0.02), offering clear interpretations and identifying significant genomic locations and brain morphological/connectivity patterns associated with SZ.

\vspace{-10pt}


\vspace{-10pt}

\end{abstract}

\section{Introduction}

\vspace{-5pt}

According to the World Health Organization (WHO), approximately 24 million individuals worldwide are affected by Schizophrenia (SZ), a condition characterized by persistent hallucinations and disruptive behavior that significantly impacts their daily life \cite{cannon2000prospective}. The symptoms of SZ often overlap with those of other psychiatric disorders, complicating both diagnosis and treatment \cite{park2023graph}. Structural magnetic resonance imaging (sMRI) is often employed to provide detailed images of brain anatomy, enabling the assessment of brain morphology and the detection of abnormalities associated with various neurological disorders, including SZ \cite{haukvik2013schizophrenia}. To gain a deeper understanding of SZ and improve diagnostic accuracy along with treatment efficacy, it is imperative to complement structural imaging data with other types of data, such as functional connectome and genomic data. Functional magnetic resonance imaging (fMRI) yields connectome data, capturing brain activity patterns and functional connections via spatial independent component analysis, crucial for identifying aberrant functional network connectivity patterns associated with SZ \cite{fu2023functional}. Single nucleotide polymorphisms (SNPs) are key genetic variations in the human genome, pivotal for understanding phenotypic traits and disorders such as SZ \cite{kim2007snp}. Each of these modalities offers unique insights into different aspects of the disorder \cite{fu2023functional}. Integrating these diverse modalities, imaging \& genomics [IMAGEN], could reveal new pathways by linking macroscopic brain differences with microscopic molecular insights into neuro-degenerative diseases including SZ \cite{mascarell2020imagen}.

Multiple existing studies have explored utilizing individual modalities or their fusion through a simple concatenation \cite{kanyal2023multi} \cite{chen2023classification} \cite{jafri2008method}. However, methods relying on feature concatenation potentially overlook crucial inter-modality relationships. Addressing this, attentional feature fusion was introduced to attentively integrate multi-modal data in a linear way \cite{dai2021attentional}. However, this may overlook complex interrelationships especially with increased modalities, potentially limiting integration efficacy. In addition to these, literature often resorts to methods like SparseCCA (SCCA) \cite{du2018fast}, Graph Neural Networks (GNN) \cite{chan2022combining}, \& Attentive DeepCCA (Att-DCCA) \cite{zhou2023attentive}, for integrating imaging genetics. Attentive DeepCCA, SparseCCA aims to find linear projections by maximizing joint correlation between 2 modalities having considerations in modeling complex \& nonlinear relationships. GNNs fuse image \& omics data through Graph Convolution (GCN), learning multi-modal relationships, but may not capture cross-modal interactions as explicitly. The vision transformer (ViT), has been employed to integrate functional and structural MRI features, providing a non-linear approach to learning attentive representations \cite{bi2023multivit}. Despite success in integrating multiple imaging modalities, a major challenge lies in integrating imaging with genomic biomarkers due to their complex nature, differing dimensionality, scale, and intricate relationships \cite{chuang2023multimodal} \cite{shaik2024gated}. Genomic biomarkers may capture molecular variations influencing brain structure and function, further complicating integration \cite{sullivan2024schizophrenia}. Addressing these challenges, we propose a novel three-way step-wise attentive integration approach that leverages the complementary information across genomics, connectome, and sMRI features to enhance SZ diagnostic accuracy. The integration order (genomics$\rightarrow$connectome$\rightarrow$sMRI) is strategically chosen based on inherent correlations between the modalities, and any change could impact the model's performance \cite{thompson2013genetics}. Primary contributions of this research include:

\begin{itemize}
    \item \textbf{Attentive Integration for Multi-modal Imaging Genomics:} Integrating genomic and connectome features attentively through Cross-modal Multi-Head Attention, prioritizing genomic and connectome interactions, and further integrating with sMRI features, selectively attending to relevant regions, and extracting complementary information from all modalities. 
    \item \textbf{Clinical Interpretability:} Ranking the top SNPs, identifying significant connections within the connectome, and highlighting SZ-specific regions in sMRIs for enhanced clinical insight.
\end{itemize} 
\vspace{-10pt}

\begin{figure}[t]
\centerline{\includegraphics[width=.95\textwidth]{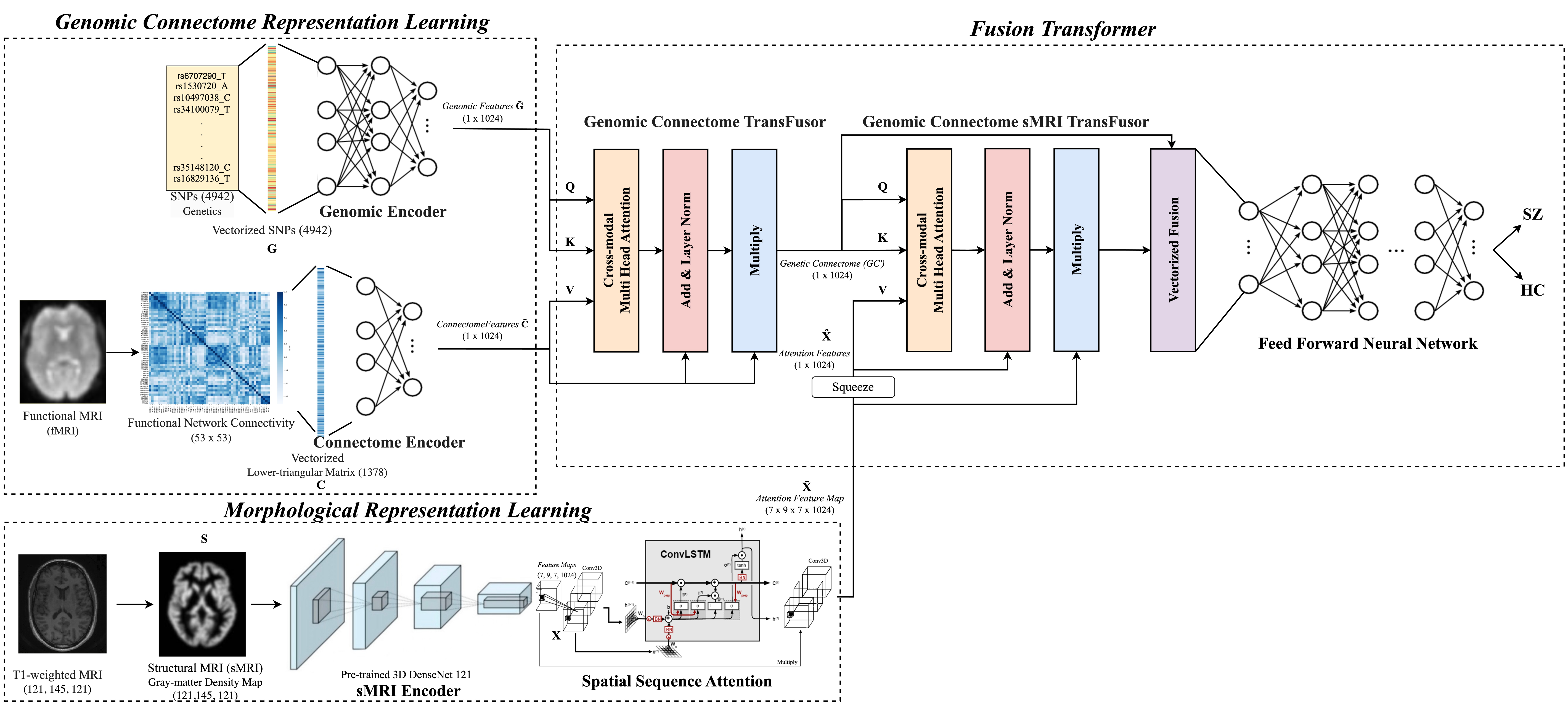}}
\caption{Architecture of Multi-modal Imaging Genomics Transformer (MIGTrans) - employing \textbf{Genomic \& Connectome Encoders} for genomic \& connectome representation learning; \textbf{Structural MRI encoder} and \textbf{Spatial Sequence Attention} for morphological representation learning; \textbf{Fusion Transformer} for multi-modal step-wise attentive integration} \label{fig:architecture}
\vspace{-5pt}
\end{figure}

\section{Multi-modal Imaging Genomics Transformer}
This study presents our Multi-modal Imaging Genomics Transformer (MIGTrans), a deep learning model that combines structural brain imaging and functional connectome with genomic data using Cross-modal Multi-Head Attention \cite{vaswani2017attention} to distinguish between individuals with SZ and healthy controls (HCs). Figure \ref{fig:architecture} illustrates MIGTrans, while subsequent sections delve into its technical modules.

\subsection{Genomic Connectome Representation Learning}
\subsubsection{Genomic Encoder}
We define the learning of genomic representations through a neural network function $\phi(\cdot)$, which takes SNP, the genomic descriptor,  $G \in \mathbb{R}^{d}$ as input and learns non-linear genomic representations $\bar{G} \in \mathbb{R}^{d'}$. This network consists of two dense layers with 2048 and 1536 units, activated by the Gaussian Error Linear Unit (GELU). Additionally, each layer is equipped with layer normalization and dropout mechanisms, ensuring robustness and preventing overfitting. The final layer outputs latent representations of the genomic features $\bar{G}$, encapsulating crucial information for discriminating SZ pathology.
\begin{equation}\label{eq:G_prime}
G^{\prime} = \text{LayerNorm}(\text{Dropout}(\Gamma(G \cdot W_1 + b_1), p_1))
\end{equation}
\vspace{-15pt}
\begin{equation}\label{eq:G_bar}
\bar{G} = \Gamma(\text{LayerNorm}(\text{Dropout}(\Gamma(G^{\prime} \cdot W_2 + b_2), p_2)) \cdot W_3 + b_3)
\end{equation}
Equations \ref{eq:G_prime} and \ref{eq:G_bar} illustrate the operations of $\phi(G)$ where $G^{\prime}$ denote intermediate genomic representations processed by neural network layers. The GELU activation $\Gamma(x) = x \cdot \frac{1}{2} \left(1 + \frac{2}{\sqrt{\pi}} \int_0^{\frac{x}{\sqrt{2}}} e^{-\tau^2} d\tau \right)$ ensures non-linearity and preserves genomic complexities \cite{hendrycks2016gaussian}. $W_1$, $W_2$, and $W_3$ are weight matrices, and $b_1$, $b_2$, and $b_3$ are bias terms applied during linear transformations. $p_1$ and $p_2$ are control dropout probabilities.
\vspace{-10pt}
\subsubsection{Connectome Encoder}
Similar to genomics, we define connectome representation learning through another neural network function $\psi(\cdot)$, which takes the functional connectome $C \in \mathbb{R}^{f}$ as input and learns non-linear connectome representations $\bar{C} \in \mathbb{R}^{d'}$. This function comprises a dense layer with 1536 units, activated by GELU, followed by layer normalization and dropout mechanisms. Subsequently, the output layer generates latent representations of the connectome, vital for capturing intricate brain connectivity patterns.  

\begin{equation}\label{eq:C_prime}
C^{\prime} = \text{LayerNorm}(\text{Dropout}(\Gamma(C \cdot W_4 + b_4), p_3))
\end{equation}
\begin{equation}\label{eq:C_bar}
\bar{C} = \Gamma(C^{\prime} \cdot W_5 + b_5))
\end{equation}
Equations \ref{eq:C_prime} and \ref{eq:C_bar} depict the processing steps of $\psi(C)$, where $C^{\prime}$ represents an intermediate connectome representation refined by the neural network layer, and $\bar{C}$ denotes the final output containing learned connectome features. $\Gamma(\cdot)$ denotes the GELU activation function. The weight matrices $W_4$ and $W_5$, along with bias terms $b_4$ and $b_5$, facilitate linear transformations essential for feature extraction. Additionally, the dropout probability $p_3$ regulates information flow, preventing overfitting, while layer normalization enhances the robustness of the network by standardizing input values.
\vspace{-5pt}
\subsection{Morphological Representation Learning}

\subsubsection{Structural MRI Encoder}
We employ Transfer Learning paradigm and use a pre-trained 3D DenseNet121 to extract morphological features from gray matter density in sMRI images. This model transforms input sMRI $S$ using a sequence of operations, including 3D convolutions, pooling, dense layers, and transition layers. The resulting output vector $X$ encapsulates the subject's brain morphological features and is passed to a spatial sequence attention (SSA) module for learning attention across spatial and channel dimensions.
\vspace{-10pt}
\subsubsection{Spatial Sequence Attention}
The SSA mechanism is tailored to capture spatial and channel dependencies within morphological feature maps $X$ \cite{shaik2022hinge}. Comprising a 3D convolutional (Conv3D) layer, a ConvLSTM layer, and another Conv3D layer, each component plays a crucial role in enhancing the feature maps \cite{shaik2024spatial}. The initial Conv3D layer extracts holistic features from the input, while the ConvLSTM, a variant of LSTM with convolutional operations, captures intricate spatial interactions and relationships across channels. By incorporating tanh and sigmoid activation functions, the ConvLSTM controls the flow of information and learns from sparse connections between input and state transitions, enabling it to learn both short and long-term dependencies within 3D feature maps \cite{shi2015convolutional}. Finally, the output of Conv3D layer projects the refined features back to the original feature space. The evolution of the cell state (${C}_t$) and hidden state (${H}_t$) at each time step $t$ is determined by input gate ($I_t$), forget gate ($F_t$), and output gate ($O_t$), considering both the current and past feature maps in the input sequence.
\begin{equation}
I_t = \sigma(W_{xi} * {X}_t + W_{hi} * {H}_{t-1} + W_{ci} * {C}_{t-1} + {b}_i)  \label{eq:it}
\end{equation}
\begin{equation}
F_t = \sigma(W_{xf} * {X}_t + W_{hf} * {H}_{t-1} + W_{cf} * {C}_{t-1} + {b}_f) \label{eq:ft}
\end{equation}
\begin{equation}
    {C}_{t} = F_t \odot {C}_{t-1} + I_t \odot \tanh(W_{xc} * {X}_t + W_{hc} * {H}_{t-1} + {b}_c) \label{eq:ct}
\end{equation}
\begin{equation}
    O_t = \sigma(W_{xo} * {X}_t + W_{ho} * {H}_{t-1} + W_{co} * {C}_{t-1} + {b}_o) {H}_{t} \label{eq:ot}
\end{equation}

\noindent The ConvLSTM operates based on fundamental equations \ref{eq:it} to \ref{eq:ot}, involving convolution ($*$) and element-wise product ($\odot$). The input gate ($I_t$) evaluates the importance of input feature maps (${X}_t$) for inclusion in the cell state (${C}_{t}$), while the forget gate ($F_t$) determines the retention of features from the previous cell state (${C}_{t-1}$). The cell state (${C}_{t}$) is updated by selectively integrating input features and forgetting unwanted ones. Subsequently, the output gate ($O_t$) regulates the exposure of the updated cell state as the output (${H}_{t}$). Through learned weight matrices ($W_{x_i}, W_{h_i}, W_{c_i}, \ldots$) and bias vectors (${b}_i, {b}_f, {b}_c, \ldots$), the ConvLSTM adapts to the specific characteristics of the data, effectively capturing spatial and channel dependencies within morphological feature maps. Finally, SSA outputs attentive sMRI features $\bar{X}$.
\vspace{-5pt}
\subsection{Fusion Transformer}
\subsubsection{Genomic Connectome TransFusor}
In this module, the learned genomic  $\bar{G}$ and connectome $\bar{C}$ features are attentively integrated through Cross-modal Multi-Head Attention (\textsl{x}-MHA) with 2 heads, emphasizing the fusion of functional connectome from fMRI with non-imaging genomic biomarkers. Genomic features are transformed and passed as Query $(Q = Linear(\bar{G}))$, Key $(K = Linear(\bar{G}))$ and connectome features are passed as Value $(V)$ to \textsl{x}-MHA. This strategic arrangement enables the model to compute attention between genomic and connectome features through a scaled dot-product mechanism, referred as Self-Attention (SA). Precisely, the attention scores are computed as the softmax of the scaled dot-product of the transformed genomic features, facilitating the selective focus on relevant interactions, and multiplying with connectome features. The resulting representation is combined with the original connectome features using layer normalization and element-wise addition. Further, the resultant features undergo element-wise multiplication with the original connectome features, yielding the final fused connectome representation $(GC^{'})$.
\vspace{-5pt}
\begin{equation}
     SA(Q, K, V) = \text{Softmax}(\frac{QK^{T}}{\sqrt{d^{k}}})V \label{eq:sa}
\end{equation}
\begin{equation}
    \text{\textsl{x}-MHA}(Q, K, V) = (\text{h}_1 \oplus ... \oplus \\ \text{h}_i) W^{o} \quad \forall \text{h}_i = \text{SA}(QW^{Q}_i, KW^{K}_i, VW^{V}_i) \label{eq:x-MHA}
\end{equation}
\begin{equation}
    GC^{'} = \bar{C} \odot \text{LayerNorm}(\bar{C} + \text{\textsl{x}-MHA}(Q, K, V)) \label{eq:gc_att}
\end{equation}
Equations \ref{eq:sa} to \ref{eq:gc_att} illustrate the mathematical operations in scaled dot-product attention and x-MHA with two heads, as well as the formulation of the genomic connectome attention features. The symbols $Q$, $K$, and $V$ represent matrices of queries, keys, and values, respectively. $d^{k}$ denotes the dimensionality of the keys, while $W^{Q}_{i}$, $W^{K}_{i}$, $W^{V}_{i}$, and $W^{o}$ are parameter matrices for linear transformation. The $\oplus$ signifies concatenation, $\odot$ indicates point-wise multiplication and the Softmax function is applied element-wise to normalize the attention scores in scaled dot-product attention. This approach is tailored to extract meaningful insights from the complex interplay between genomic and connectome data, enabling the model to effectively capture intricate dependencies specific to SZ.
\vspace{-10pt}
\subsubsection{Genomic Connectome sMRI TransFusor}
Similar to the previous module, Genomic Connectome integration with sMRI is facilitated through \textsl{x}-MHA mechanism with 2 heads. This module aims to capture cross-modal relationships between imaging and non-imaging biomarkers. The genomic connectome features ($GC^{'}$) are passed as Query $\widetilde{Q}$ and Key $\widetilde{K}$, while the squeezed sMRI attention features $(\hat{X})$ are passed as Value $\widetilde{V}$ to the \textsl{x}-MHA. This strategic setup enables the model to selectively attend to relevant regions in the sMRI features based on the genomic connectome attention features, thereby incorporating complementary information between modalities. The attention-weighted representations are fused with the original sMRI features through a process of layer normalization and element-wise addition. This fusion step ensures that the model can adaptively combine information from multi modalities while preserving the structural characteristics of the sMRI. Subsequently, the fused representations undergo further vectorized fusion through dense layers. This helps retain cross-modality features alongside fused representations, thereby improving the model's generalization performance.
\vspace{-5pt}
\begin{equation}
    GCS^{'} = \bar{X} \odot \text{LayerNorm}(\hat{X} + \text{\textsl{x}-MHA}(\widetilde{Q}, \widetilde{K}, \widetilde{V})) \label{eq:gci_att}
\end{equation}
Equation \ref{eq:gci_att} elucidate the post-processing fusion after \textsl{x}-MHA, whose operations were illustrated in equations \ref{eq:sa} and \ref{eq:x-MHA}. The Feed Forward Neural Network acts as a classification head and utilizes the fused feature representations $GCS^{'}$ to predict the target labels $\hat{y} \in \{SZ, HC\}$. It consists of two dense layers with 512, 256 units each, leveraging GELU activation to extract intricate features and introduce non-linearity, enhancing the model's representational capacity. Employing layer normalization and a dropout mitigates overfitting, fostering improved model generalization by preventing excessive reliance on specific features. All the modules discussed above were trained end-to-end to minimize the cross-entropy loss, $\mathcal{L}(y, \hat{y}) = -\frac{1}{N} \sum_{i=1}^{N} \left( y_i \log(\hat{y}_i) + (1 - y_i) \log(1 - \hat{y}_i) \right)$, using the Adam optimizer for gradient optimization. The cross-entropy loss function \( \mathcal{L} \) measures the dissimilarity between true labels \( y \) and predicted probabilities \( \hat{y} \). 

\vspace{-5pt}
\section{Experiments \& Results}
\vspace{-5pt}
\subsection{Dataset, Pre-processing \& Experimental Setup}
\vspace{-5pt}
To prove the effectiveness of the proposed MIGTrans, we employed a subset of the Function Biomedical Informatics Research Network (FBIRN) dataset, containing sMRI, fMRI, and SNP data from 186 participants, including 82 SZ and 104 HC subjects \cite{keator2016function}. In our work, we select a subset of 4942 SNPs (one-hot encoded) related to SZ as identified by the psychiatric genomics consortium, extract the lower triangular matrix from $(53 \times 53)$ FNC matrix generated using Neuromark Atlas \cite{du2020neuromark}, resulting in 1378 unique connections. Additionally, each sMRI is downsampled to $(121 \times 145 \times 121)$ dimensions \& converted to gray matter density map using SPM12 \cite{kurth201512}. We employ 5-fold cross validation to evaluate the effectiveness of proposed model. Throughout our experiments, various hyperparameter values were explored, including learning rates (0.005 to 0.05), dropout rates (0.1 to 0.5), and regularization rates (0.001 to 0.1). Selected values to optimize model performance: initial learning rate (0.001), dropout ($p_1$=0.5, $p_2,p_3$=0.3), batch size (16), training epochs (100), L2 weight regularization (0.005) and L1\_L2 bias regularization (0.005). 
\vspace{-10pt}
\subsection{Quantitative Evaluation}
\begin{table}[b]
\vspace{-10pt}
\caption{Classification performance comparison of proposed model with baselines}\label{tab:results}
\vspace{-5pt}
\centering
\begin{tabular}{|l|c|c|c|c|}
\hline
\textbf{Model} &\textbf{ Accuracy} (\%) & \textbf{Precision} (\%) & \textbf{Recall} (\%) & \textbf{$F_1$ Score} (\%) \\
\hline
Genomic SANet & 63.64($\pm$0.04)& 64.48($\pm$0.04)& 60.77($\pm$0.03)& 59.28($\pm$0.03)\\
Connectome SANet & 82.33($\pm$0.04)& 81.87($\pm$0.03)& 82.04($\pm$0.03)& 81.88($\pm$0.03) \\
sMRI SSANet & 75.71($\pm$0.06)& 78.59($\pm$0.04)& 72.81($\pm$0.08)& 72.12($\pm$0.09) \\ \hline
GC Concat \cite{venugopalan2021multimodal} & 82.71($\pm$0.05)& 82.99($\pm$0.04)& 81.76($\pm$0.04)& 82.04($\pm$0.04) \\
GC AFF \cite{dai2021attentional} & 83.83($\pm$0.04)& 83.54($\pm$0.03)& 83.67($\pm$0.04)& 83.38($\pm$ 0.04) \\
{GCTrans} & {84.82($\pm$0.05)}& {84.45($\pm$0.05)}& {84.56($\pm$0.05)}& {84.45($\pm$0.05)} \\ \hline
GCS SCCA \cite{du2018fast} & 84.29($\pm$0.07) & 83.86($\pm$0.04) & 83.99($\pm$0.04) & 83.92 ($\pm$0.05) \\
GCS Att-DCCA \cite{zhou2023attentive} & 84.31($\pm$0.05) & 84.56($\pm$0.04) & 84.02($\pm$0.04) & 83.88 ($\pm$0.05) \\
GCS Concat \cite{venugopalan2021multimodal} & 84.32($\pm$0.06)& 84.55($\pm$0.06)& 83.46($\pm$0.05)& 83.79($\pm$0.05) \\
GCS AFF \cite{dai2021attentional} & 84.30($\pm$0.05)& 84.48($\pm$0.06)& 83.53($\pm$0.04)& 83.77($\pm$0.05) \\
\textbf{Proposed} & \textbf{86.05($\pm$0.02)}& \textbf{87.32($\pm$0.03)}& \textbf{84.60($\pm$0.02)}& \textbf{85.19($\pm$0.02)} \\
\hline
\end{tabular}
\vspace{-10pt}
\end{table}
We conducted initial experiments to check the performance of single-modality approach using attention mechanism. We designed Self-Attention Networks for genomics (Genomic SANet) and connectome (Connectome SANet), and Spatial Sequence Attention Network (SSANet) for sMRI features. For multi-modal approaches, we explored various fusion strategies such as simple concatenation (Concat), Attentive Feature Fusion (AFF), SCCA, Att-DCCA for comparison with our proposed transformer-based method (Trans). As an ablation study, we also report the performance of the first-step fusion with genomics, connectome (GC) and the follow-up fusion with sMRI (GCS). Table \ref{tab:results} summarizes the classification performance of our proposed method and the baselines. The connectome alone performed best among single-modal features with 82.33\%, supporting the SZ clinical finding \cite{kanyal2023multi}. However, this is slightly lower than the 82.71\% achieved by simply combining genomic and connectome attentive features.From this, it is clearly evident that multi-modal features are superior to when they are considered in isolation.  In terms of multi-modal feature integration, our proposed Trans method showed significant improvement in all metrics for two-way (GC) fusion when in comparison with concatenation and AFF, indicating 2.11\%, 0.99\% improvement. This trend continued even in three-way (GCS) fusion with 1.72\% 1.73\%, 1.74\% 1.75\% improvement when compared to SCCA, AFF, Att-DCCA and Concat in terms of accuracy. This highlights the benefit of our step-wise attentive integration of imaging with genomic biomarkers through cross-modal multi-head attention. Additionally, it is noteworthy that AFF showed a 0.02\% lower accuracy compared to concatenation in three-way GCS fusion, indicating its struggle to maintain superiority as the number of modalities to integrate increases, given its reliance on linear fusion.
\vspace{-5pt}
\begin{figure}[t]
\centerline{\includegraphics[width=.9\textwidth]{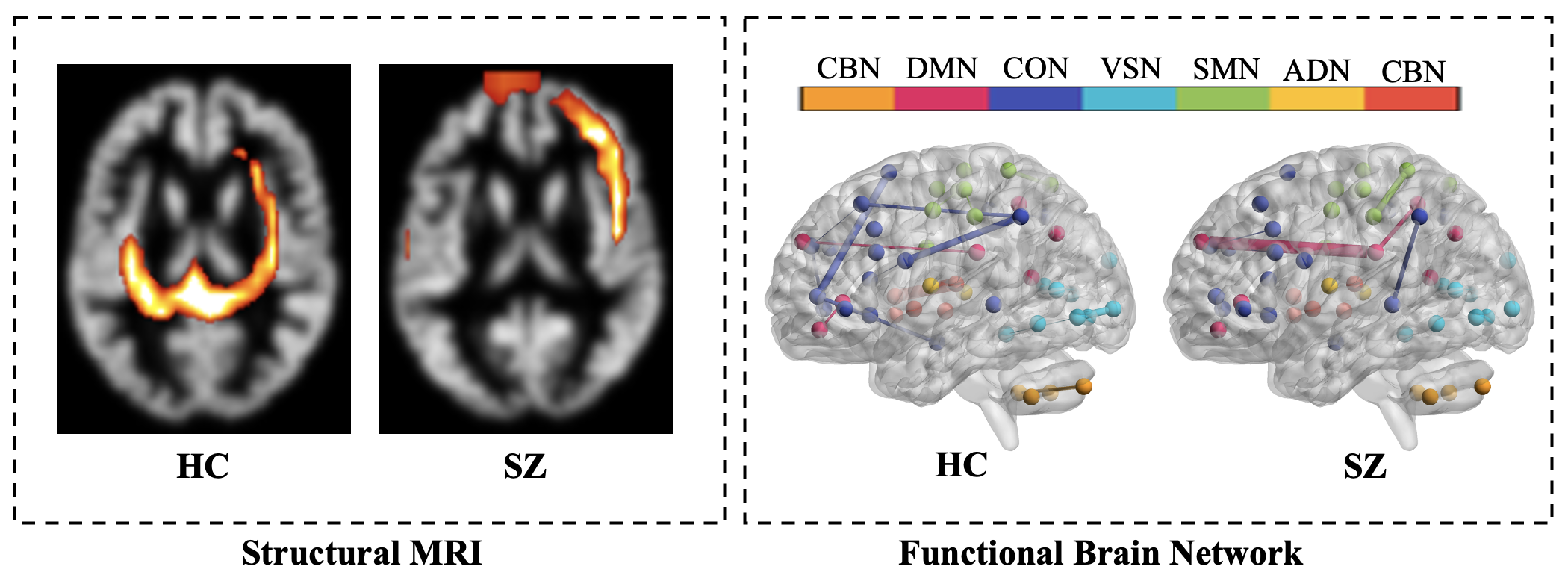}}
\vspace{-5pt}
\caption{\textbf{Attention maps of sMRI} highlighting the frontal and temporal lobes in SZ samples, contrasting with thalamic regions in HC samples; \textbf{Functional brain networks} highlighting important connections in default mode (DMN) and cognitive control (CON) specific to SZ and HC compared with other subcortical (SCN), auditory (ADN), sensorimotor (SMN), visual (VSN), and cerebellum (CBN) networks.} \label{fig:att-maps}
\vspace{-10pt}
\end{figure}

\vspace{-5pt}
\subsection{Qualitative Evaluation}
\vspace{-5pt}
To provide interpretation and clinical validation of our approach, we presented the attention maps of sMRI and important connections in functional connectome generated based on attention scores in Fig. \ref{fig:att-maps}.  For sMRI, attention scores from genome connectome sMRI TransFusor were multiplied with SSA Attention maps and visualized by GradCAM++. These maps identify primary regions highlighted in the frontal and temporal lobes in SZ samples, contrasting with thalamic regions in HC samples. This aligns with clinical observations indicating that individuals with SZ experience the most significant reductions in thalamic volumes and thickness within the frontal and temporal lobes \cite{haukvik2013schizophrenia}. For functional brain network, we applied thresholding on FNC Attention scores to find top connectome features and observed denser connectivity patterns in HC subjects compared to individuals with SZ. In particular, the Default Mode Network (DMN) and Cognitive Control Network (CON) were revealed as significant interactions for SZ classification, aligning with previous study \cite{identification2010kim}. Moreover, our approach successfully pinpointed the top three SZ-associated SNPs (\textit{rs7643366, rs396911, rs13429970}), consistent with a genome-wide association study \cite{ebi-gwas} \cite{kato2023genetic}. These results emphasize the discriminative potential of the attentive integration mechanisms in our MIGTrans.

\vspace{-7pt}
\section{Conclusion}
\vspace{-5pt}
The Multi-modal Imaging Genomics Transformer (MIGTrans) integrates genomics, connectome, and structural imaging features through a novel three-way step-wise attentive integration, offering a comprehensive approach to capturing structural/functional brain abnormalities and genetic factors associated with schizophrenia. This approach surpasses the performance of baseline methods, demonstrating its effectiveness in enhancing clinical decision-making and improving diagnostic accuracy. With its ability to leverage complementary information from diverse data modalities, it identifies significant genomic locations, captures brain connectivity patterns, and highlights spatial regions specific to schizophrenia. MIGTrans holds promise for advancing our understanding of schizophrenia and facilitating more personalized therapeutic interventions in clinical settings.

\bibliographystyle{splncs04}
\bibliography{refs}
\end{document}